\pdfoutput=1

\documentclass[11pt]{article}

\usepackage[]{ACL2023}

\usepackage{times}
\usepackage{latexsym}
\usepackage{amssymb}
\usepackage{pifont}
\usepackage{graphicx}
\usepackage{verbatim}
\usepackage{xcolor}
\usepackage{array}
\usepackage{makecell}
\usepackage{booktabs}
\usepackage{subcaption}
\usepackage{microtype}
\usepackage{multirow}
\usepackage{amsmath}

\usepackage[T1]{fontenc}

\usepackage[utf8]{inputenc}

\usepackage{microtype}

\usepackage{inconsolata}

\title{A Simple and Flexible Modeling for Mental Disorder Detection\\by Learning from Clinical Questionnaires}

\author{Hoyun Song \hspace{6mm}
Jisu Shin\hspace{6mm}
Huije Lee \hspace{6mm}
Jong C. Park$\thanks{\hspace{2mm}Corresponding author}$ \\
School of Computing \\
Korea Advanced Institute of Science and Technology\\
\texttt{\{hysong1991,jisu.shin,angiquer,jongpark\}@kaist.ac.kr}\\
}

\begin{document}
\maketitle

\begin{abstract}
Social media is one of the most highly sought resources for analyzing characteristics of the language by its users. In particular, many researchers utilized various linguistic features of mental health problems from social media. However, existing approaches to detecting mental disorders face critical challenges, such as the scarcity of high-quality data or the trade-off between addressing the complexity of models and presenting interpretable results grounded in expert domain knowledge. To address these challenges, we design a simple but flexible model that preserves domain-based interpretability. We propose a novel approach that captures the semantic meanings directly from the text and compares them to symptom-related descriptions. Experimental results demonstrate that our model outperforms relevant baselines on various mental disorder detection tasks. Our detailed analysis shows that the proposed model is effective at leveraging domain knowledge, transferable to other mental disorders, and providing interpretable detection results.
\end{abstract}
\section{Introduction}
\label{sec:introduction}
Mental health problems, a significant challenge in public healthcare, are usually accompanied by distinct symptoms, such as loss of interest or appetite, depressed moods, or excessive anxiety.
As these symptoms can often be expressed over social media, detecting mental health conditions using social media text has been studied extensively \cite{yates2017depression, coppersmith2018natural, matero2019suicide, murarka2020detection, harrigian2021state,
jiang2021automatic, nguyen2022improving}. 
Such approaches could give rise to a monitoring system that provides clinical experts with information about possible mental crises.

To automatically identify mental health problems, traditional approaches focus on finding linguistic patterns and styles from the language of psychiatric patients.
Utilizing these features, statistical models can explain the correlation between linguistic factors and mental illnesses.
However, these approaches suffer from increased complexity of models, necessitating pipelines of steps, from engineering features to producing results. 
By contrast, more recent works have employed strong pre-trained models, which allow a direct use of raw data and simplify model development \cite{matero2019suicide, jiang2020detection}.
While such end-to-end approaches may be effective at achieving higher performance, they often lack domain-based interpretation, which is essential for decision-support systems \cite{mullenbach2018explainable}.
Hence, there is a trade-off between providing interpretable predictions based on domain knowledge and the simplicity of the models.


The lack of a sufficient sample size for high-quality data is another challenge in the clinical domain \cite{de2017gender, harrigian2020models}.
Despite the availability of diverse datasets and methods for detecting mental disorders, most of them aim primarily at identifying only clinical depression.
To tackle such a problem, recent studies have focused on developing transferable linguistic features that can be used for the detection of various mental disorders \cite{aich2022really, uban2022multi}.
However, the linguistic features that are trained on a particular dataset may not be fully transferable to a different task \cite{ernala2019methodological,harrigian2020models}.

Others utilized symptom-related features that are more common properties of psychiatric patients, resulting in generalizability of depression detection \cite{nguyen2022improving}.
Despite this improvement, however, their approach still faces challenges because they rely on pipelined methods using manually-defined symptom patterns.
Such symptom patterns for depression detection lack flexibility as they cannot be easily adapted to other mental disorders.
In addition, the pipeline approach with symptom extraction is quite complex to implement.
It involves multiple steps, designing symptom patterns, training a symptom identification model, and detecting depression using the identified symptom patterns.

To address these challenges, we propose to design a simple and more flexible approach that also preserves interpretability.
We are motivated by the process that humans use to quickly learn related features, often by reading just a single explanation.
For example, when people are reading depression questionnaires, they readily understand the questions and learn about symptoms that are related to depression, allowing them to self-diagnose their levels of depression.

To this end, we employ the siamese network \cite{koch2015siamese}, which captures the semantic meaning of the text inputs and compares them directly to symptom-related descriptions.
This process is simple since they find symptom-related clues directly from the input, rather than relying on hand-engineered features or intermediate models.
Our proposed model, Multi-Head Siamese network (MHS), can be easily adapted to other mental illness domains by simply replacing the symptom-related descriptions.
In addition, our model is designed to capture the distinct features of each symptom using multiple heads.
By examining the learned weights of each symptom head, our model gives rise to human-understandable interpretations.

We evaluate the performance of our model, detecting texts containing mental health problems on four mental disorders. 
Furthermore, the detailed analysis of the proposed model shows its efficiency in utilizing symptom-related knowledge, its ability to be applied to different mental disorders, and its interpretable reasoning for detected results.

\section{Related Work}
\label{sec:relatedwork}
Social media are commonly used for mental health research because of the ease of access to various aspects of human behavior studies.
Similarly to other NLP domains, pre-trained language models, such as BERT \cite{devlin2019bert}, are widely used for identifying mental health problems \cite{matero2019suicide, jiang2020detection, murarka2020detection, dinu2021automatic}.

Others have presented interpretable detection methods for the mental health domain based on linguistic features \cite{song2018feature, uban2021understanding}. 
Various efforts have also been made to study such linguistic features accompanying mental illness, such as differences in word usage \cite{tadesse2019detection, jiang2020detection, dinu2021automatic}, or in syntactic features \cite{kayi2018predictive, ireland2018within, yang2020big}.
Some studies address the differences between sentiments or emotional aspects \cite{preoctiuc2015mental, gamaarachchige2019multi, allen2019convsent,  wang2021learning}, or differences in topics \cite{tadesse2019detection, kulkarni2021cluster}.

The linguistic features are also used for transferable methods across other mental disorders \cite{aich2022really, uban2022multi}, focusing on the fact that a large number of studies have been done primarily on depression \cite{de2013predicting, yates2017depression, eichstaedt2018facebook, song2018feature, tadesse2019detection, yang2020big, nguyen2022improving}, compared to other disorders, such as anxiety disorder \cite{ireland2018within}, anorexia \cite{uban2021understanding}, or schizophrenia \cite{kayi2018predictive}.
However, such linguistic features do not generalize well to new user groups.
For example, \citet{de2017gender}, \citet{loveys2018cross}, and \citet{pendse2019cross} found that the linguistic styles may vary to their backgrounds. 
In addition, \citet{harrigian2020models} found that a model trained on a particular dataset does not always generalize to others.
To handle such a generalization problem, \citet{nguyen2022improving} and  \citet{zhang2022symptom} focused on the shared and general properties (i.e., symptoms) of a mental health problem.
However, unlike ours, which captures the symptom features directly from raw data, these methods require additional steps for learning symptom-related features.

In this paper, we use the siamese network \cite{koch2015siamese}, based on one-shot learning, exploited recently for simple networks \cite{chen2021exploring, zhu2021task}.
We utilize the symptom descriptions sourced from DSM-5 \cite{american2013diagnostic} to make our model learn symptom-related knowledge.

\section{Methodology}
\label{sec:method}
In this section, we introduce our simple but flexible modeling for leveraging clinical questionnaires.
Our model aims to detect texts with mental illness episodes based on the presence of symptom-related features just by a single component.

\begin{figure*}[ht]
\centering
\includegraphics[scale=0.4]{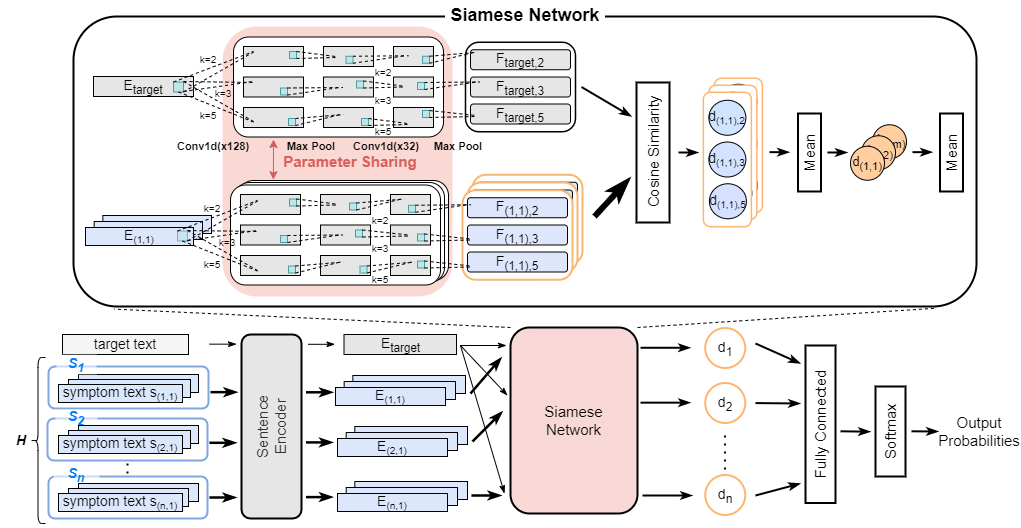}
\caption{The model architecture of Multi-Head Siamese network (MHS).
$S_i$ indicates a head of a symptom that contains the symptom-related descriptions $s_{(i, j)}$, and $d_i$ indicates a distance value computed by cosine similarity between the target text and the descriptions.
MHS compares the contextualized embeddings of the target text and symptom and predicts the probability of mental illness based on context similarity.
}
\label{fig:model}
\end{figure*}

An overview of our network is shown in Figure~\ref{fig:model}.
We designed our model based on the siamese network \cite{koch2015siamese}.
As with the original siamese neural network, our model also contains a single feature extractor with shared parameters.
The extractor directly obtains features from contextualized embeddings generated by sentence encoders.
Then, employing the similarity function, we compare the similarity to see the presence of symptom-related features from the target text.
In addition, we apply multi-headed learning to the original siamese network, repeating the comparison process for each distinct symptom.
We describe the detailed structure in the following subsections.

\subsection{Model Structure}
\label{subsec:model}
Our model, the Multi-Head Siamese network (MHS), is an end-to-end model that takes raw input texts and produces the final result without the need for manual feature engineering.
MHS is designed to take two types of inputs, the target text to be classified and descriptions of symptoms.
The descriptions are grouped for each symptom, and each symptom group is the input for the corresponding symptom head.
For example, assuming that we have $n$ symptoms for discriminating against mental disorder, we build a set of $n$ heads ($H$) from $S_1$ to $S_n$ for the detection model as follows:
\begin{equation}
    H = \{S_1, S_2, ..., S_n\}
\end{equation}
\noindent Each head $S$ represents discrete symptoms, containing a number of descriptions and questions regarding the corresponding symptom.
For example, if $S_i$ has $m$ sentences describing the symptom, we have a set $S_{i}$ of questions:
\begin{equation}
    S_{i}= \{s_{(i,1)}, s_{(i,2)}, ..., s_{(i,m)}\}
\end{equation}

\noindent With a given input of the target sentence, our model obtains embedding vectors ($E_{target}$) by employing pre-trained sentence encoders, such as BERT or RoBERTa.
We also get symptom embeddings by encoding all sentences from all heads ($H$).

Our siamese network employs a multi-channel convolutional neural network (CNN) for feature learning.
We apply three channels for convolution layers, whose kernel sizes are 2, 3, and 5.
Thus, our model is designed to capture informative clues with the window sizes of 2, 3, and 5 from texts.
Each channel contains two convolutional layers and two max-pooling layers.
The final convolutional layer is flattened into a single embedding vector. 
As a result, we obtain three feature embedding vectors ($F_{target,k}$) with $k={2,3,5}$ from the target text:
\begin{equation}
    F_{target,k} = Conv1d_k(E_{target})
\end{equation}
\noindent Through the same process, we also obtain feature embedding vectors from symptom texts from the $i^{th}$ head and $j^{th}$ sentence as follows:
\begin{equation}
    F_{(i,j),k} = Conv1d_k(E_{(i,j)})
\end{equation}

We compute the distances ($d$) between the target feature vector ($F_{target,k}$) and a symptom-sentence vector ($F_{(i,j),k}$) using cosine similarity, ranging from $[-1,1]$.
We calculate a single distance value by taking the average of $K$ distance values, where $K$ represents the number of channels:
\begin{equation}
    sim({\bf x},{\bf y}) = {\frac{{\bf x} {\bf y}}{\|{\bf x}\| \|{\bf y}\|}}  
\end{equation}
\begin{equation}
    d_{(i,j)} = \frac{1}{K}\sum_{k}^{}sim(F_{target,k}, F_{(i,j),k})
\end{equation}



\noindent Finally, when there are distance values for all sentences, they are averaged to yield the distance value of the $i^{th}$ head ($d_{i}$):
\begin{equation}
    d_{i} = \frac{1}{m}\sum_{j=1}^{m}d_{(i,j)}
\end{equation}

\noindent To regularize the results, we choose to use averaging as an aggregation function for the distance values.

We iterate this process over the number of heads ($n$).
After the siamese network step, all distance values ($d_{i}$) are stacked into a $1\times{n}$ vector ($D$). 
By applying the fully connected layer, the distance vector is reduced into a two-dimensional vector $o$, which is an output probability of classifying mental illness: 
\begin{equation}
    f : \mathbb{R}^n \rightarrow \mathbb{R}^2
\end{equation}
\begin{equation}
    o = f(D) =  W^T \cdot D + b
\end{equation}

\noindent By analyzing the weights ($W$) and distance values ($D$) of the fully connected layer, we can examine which symptoms are activated as important information when classifying the related mental disorder. Further details are discussed in Section \ref{subsec:interpretation}.
The implementation code and symptom-sentences are made publicly available\footnote{https://github.com/HoyunSong/acl23-multi-head-siamese-mental-illness}.


\subsection{Symptom Descriptions}
\label{subsec:symptom}

\begin{table}[ht]
\centering
\scriptsize
  \begin{tabular}{c|l}
    \hline
        \multicolumn{1}{c|}{\bf\makecell[c]{Mental\\Disorders}} & \multicolumn{1}{c}{\bf Diagnostic Criteria from DSM-5}  \\
    \hline
        \multicolumn{1}{c|}{\makecell[c]{Major \\Depressive\\Disorder\\(D0-D8)}} & 
        \makecell[l]{D0. Depressed mood most of the day\\
        D1. Diminished interest or pleasure\\
        D2. Sleep disorders (insomnia or hypersomnia)\\
        D3. Changes in weight or appetite when not dieting\\
        D4. Fatigue or loss of energy\\
        D5. Feeling worthlessness or guilty\\
        D6. Diminished ability to think or concentrate\\
        D7. A slowing down of thought and a reduction of\\\hspace{3mm}physical movement\\
        D8. Recurrent thoughts of death and suicidal ideation\\
        } \\

    \hline
        \multicolumn{1}{c|}{\makecell[c]{Bipolar\\Disorder\\(D0-D8,\\M0-M7)}} & 
        \makecell[l]{\bf Major Depressive Episode\\\hspace{2mm}D0-D8: Same as major depressive disorder\\
        \bf Manic Episode\\
        \hspace{2mm}M0. A distinct period of persistently elevated\\\hspace{5mm}or expansive mood\\
        \hspace{2mm}M1. Increase in goal-directed activity\\
        \hspace{2mm}M2. Inflated self-esteem or grandiosity\\
        \hspace{2mm}M3. Decreased need for sleep\\
        \hspace{2mm}M4. More talkative than usual\\
        \hspace{2mm}M5. Flight of ideas\\
        \hspace{2mm}M6. Distractibility\\
        \hspace{2mm}M7. Activities that have a high potential for\\\hspace{5mm}painful consequences\\
        } \\
    \hline
        \multicolumn{1}{c|}{\makecell[c]{Generalized\\Anxiety\\Disorder\\(A0-A6)}} & 
        \makecell[l]{A0. Excessive anxiety and worry more than 6 months\\
        A1. Difficult to control the worry\\
        The anxiety and worry are associated with followings:\\
        \hspace{2mm}A2. Irritability\\
        \hspace{2mm}A3. Being easily fatigued\\
        \hspace{2mm}A4. Sleep disturbance\\
        \hspace{2mm}A5. Difficulty concentrating or mind going blank\\
        \hspace{2mm}A6. Muscle tension\\
        } \\
    \hline
        \multicolumn{1}{c|}{\makecell[c]{Borderline\\Personality\\Disorder\\(B0-B8)}} & 
        \makecell[l]{B0. Interpersonal relationships alternating between\\\hspace{3mm}idealization and devaluation\\
        B1. Recurrent suicidal or self-mutilating behavior\\
        B2. Identity disturbance\\
        B3. Affective instability\\
        B4. Inappropriate anger or difficulty controlling anger\\
        B5. Transient, stress-related paranoid ideation\\\hspace{3mm}or severe dissociative symptom.\\
        B6. Impulsive behaviors that are potentially\\\hspace{3mm}self-damaging\\
        B7. Frantic efforts to avoid abandonment\\
        B8. Chronic feelings of emptiness\\
        } \\
    \hline
  \end{tabular}
  \caption{\label{tab:symptoms} A summary of diagnostic criteria for each mental disorder, sourced from DSM-5.}
  \vspace{-0.2in}
\end{table}

In the present study, we focus on four mental disorders: major depressive disorder (MDD), bipolar disorder, generalized anxiety disorder (GAD), and borderline personality disorder (BPD).
As summarized in Table \ref{tab:symptoms}, we compiled the diagnostic criteria for each mental disorder, sourced from DSM-5. We constructed heads based on the list of symptoms.
For example, in the case of MDD, there are a total of 9 symptoms (D0-D8), so when constructing a model detecting depressive symptoms, there will be a total of 9 heads ($n(H_{MDD})=9$). As for bipolar disorder, symptoms can be divided into depressive episodes (D0-D8) and manic episodes (M0-M7), with a total of 17 heads.
The depressive episodes of bipolar disorder are the same as those of MDD.

Each head includes a description of diagnostic criteria and questions from self-tests corresponding to each symptom.
As a result, each head contains two or more sentences ($n(S) \ge 2$).
In the case of more than two related questions for a symptom, the corresponding head contains more than two sentences.

We collected the questions from the publicly available self-tests\footnote{MDD (www.psycom.net/depression-test/),\\
Bipolar (www.psycom.net/bipolar-disorder-symptoms/),\\
GAD (www.psycom.net/anxiety-test), and\\
BPD (www.psycom.net/borderline-personality-test/)}.
The process was conducted under the guidance of a psychology researcher.
The complete list of collected sentences for each head is shown in Appendix~\ref{appendix:symptom}.
Our model can easily transfer to other mental disorders by just replacing symptom descriptions, as evidenced by the findings in Section~\ref{subsec:cross-domain}.

\section{Experiments}

\subsection{Dataset and Evaluation}
\label{subsec:dataset}

In order to evaluate our model, we constructed four datasets to detect possible mental disorder episodes. We sampled posts from Reddit\footnote{https://files.pushshift.io/reddit/}, which is one of the largest online communities. 
Each sample is a concatenation of a title and a body from a post.
Each dataset contains two groups of Reddit posts. 
One includes the posts collected from mental disorder-related subreddits as a text containing the mental illness contents, and the other is from random subreddits as a clean text. 
The detailed statistics of each group is shown in Table \ref{tab:data}. 
We performed preprocessing by discarding posts containing URLs or individually identifiable information, and posts shorter than ten words (i.e., tokens). 
We only retained posts in English; otherwise, they are discarded. 

We conducted four tasks, employing these collected datasets, discriminating texts sourced from mental disorder-related subreddits out of non-mental illness texts.
The details of each task are as follows: MDD detection (\textit{r/depression}+random), Bipolar disorder detection (\textit{r/bipolar}+random), GAD detection (\textit{r/anxiety}+random), and BPD detection (\textit{r/bpd}+random).

To compare our model with baseline models with respect to classification performance, we report results using standard metrics, Accuracy (Acc.), F1 score (F1) for the mental illness group, and Area Under the Curve (AUC).
The performance measure is reported by five-fold cross-validation, and each repetition is trained on six different seeds.
We averaged after 30 runs (5×6) to get the final result.

\subsection{Baselines and Experimental Setup}
\label{subsec:baselines}

\begin{table}[t]
\centering
\scriptsize
  \begin{tabular}{l|rrrr}
    \hline
    \bf Subreddit & \bf \#samples & \bf sent. & \bf tok. & \bf vocab. \\
    \hline
    r/depression & 11,416 & 9.5 & 143.1 & 43,766 \\
    r/bipolar & 10,941 & 10.5 & 157.1 & 54,426 \\
    r/anxiety & 11,471 & 9.7 & 159.8 & 51,936 \\
    r/bpd & 10,979 & 11.8 & 187.5 & 53,741 \\
    Random &  40,570 & 8.8 & 123.0 & 198,988 \\
    \hline
    Total &  85,377 & 9.6 & 133.6 & 229,309 \\
    \hline
  \end{tabular}
  \caption{\label{tab:data} The number of samples, average numbers of sentences and tokens, and the vocabulary size.}
  \vspace{-0.1in}
\end{table}
In this subsection, we describe models and implementation details for experiments.
More experimental details are shown in Appendix~\ref{appendix:experimental}.

\par \textbf{1) Traditional Models} We implemented two feature-based classifiers, a support vector machine (SVM) and a random forest (RF), with two versions: \textbf{BoW}, employing lexical features only \cite{tadesse2019detection, jiang2020detection}, and \textbf{Feature}, adding sentimental and syntactic features \cite{allen2019convsent, yang2020big, wang2021learning}.
\textbf{2) BERT} \cite{devlin2019bert} is one of the most well-known baseline models using contextualized embeddings \cite{jiang2020detection, matero2019suicide}. 
\textbf{3) XLNet} \cite{yang2019xlnet} is another strong baseline with a pre-trained language model \cite{dinu2021automatic}. 
\textbf{4) RoBERTa} \cite{liu2019roberta} is a robustly optimized BERT and one of the most solid baselines in natural language classification \cite{dinu2021automatic, murarka2020detection}. 
\textbf{5) GPT-2} \cite{radford2019language} is a strong few-shot learner with a large Transformer-based language model.
\textbf{6) PHQ9} \cite{nguyen2022improving} is a depression detection model constrained by the presence of PHQ9 symptoms.

We implemented our models using PyTorch and fine-tuned our models on one 24GB Nvidia RTX-3090 GPU, taking about 13 minutes for each epoch. The batch size and embedding size of all models are 8 and 512, respectively, and are fine-tuned over five epochs. We truncated each post at 512 tokens for all models. For each model, we manually fine-tuned the learning rates, choosing one out of \{1e-5, 2e-5, 1e-6, 2e-6\} that shows the best F1 score. We report the average results over 30 runs (five-fold cross-validations are trained on six different seeds) for the same pre-trained checkpoint.

\subsection{Experimental Results}
\label{subsec:results}

\begin{table*}[t]
\scriptsize
\centering
\setlength\tabcolsep{4.4pt}
\setlength\extrarowheight{2pt}
\begin{tabular}{lllclllclllclllc}
\hline
\multicolumn{1}{c}{\multirow{2}{*}{Model}} & \multicolumn{3}{c}{\bf MDD}                                                            &  & \multicolumn{3}{c}{\bf Bipolar}                                                        &  & \multicolumn{3}{c}{\bf GAD}                                                            &  & \multicolumn{3}{c}{\bf BPD}                                                            \\ \cline{2-4} \cline{6-8} \cline{10-12} \cline{14-16} 
\multicolumn{1}{c}{}                       & \multicolumn{1}{c}{Acc.} & \multicolumn{1}{c}{F1 ($\pm$)} & \multicolumn{1}{c}{AUC} &  & \multicolumn{1}{c}{Acc.} & \multicolumn{1}{c}{F1 ($\pm$)} & \multicolumn{1}{c}{AUC} &  & \multicolumn{1}{c}{Acc.} & \multicolumn{1}{c}{F1 ($\pm$)} & \multicolumn{1}{c}{AUC} &  & \multicolumn{1}{c}{Acc.} & \multicolumn{1}{c}{F1 ($\pm$)} & \multicolumn{1}{c}{AUC} \\ \hline
RF-BoW        & 89.9 & 73.7 (0.34) & 80.4 & & 90.9 & 75.8 (0.37) & 81.1 & & 91.7 & 76.3 (0.41) & 81.7 & & 90.3 & 73.2 (0.35) & 79.8\\
SVM-Bow       & 91.2 & 78.0 (0.89) & 83.6 & & 90.2 & 78.2 (0.84) & 81.4 & & 92.9 & 83.3 (0.84) & 88.5 & & 93.4 & 83.6 (0.67) & 88.9\\
RF-Feature    & 89.6 & 72.9 (0.54) & 79.8 & & 91.1 & 76.2 (0.54) & 81.4 & & 91.8 & 79.2 (0.77) & 83.7 & & 90.4 & 73.5 (0.45) & 80.0 \\
SVM-Feature   & 92.2 & 81.5 (0.59) & 86.6 & & 93.3 & 83.6 (0.77) & 87.5 & & 94.3 & 86.7 (0.81) & 90.0 & & 93.6 & 83.6 (0.41) & 88.6\\ \hline
GPT-2         & 94.6 & 88.0 (0.51) & 92.6 & & 95.3 & 88.9 (0.63) & 92.4 & & 95.7 & 90.2 (0.35) & 93.5 & & 95.6 & 89.7 (0.49) & 93.4 \\
XLNet         & 94.4 & 87.9 (0.40) & 92.1 & & 95.2 & 88.8 (0.43) & 92.4 & & 95.7 & 89.8 (0.26) & 93.2 & & 95.6 & 89.4 (0.43) & 92.9\\\hline
BERT          & 94.2 & 87.3 (0.41) & 92.4 & & 95.0 & 88.1 (0.56) & 91.3 & & 95.3 & 88.5 (0.61) & 91.9 & & 95.0 & 88.9 (0.55) & 93.2 \\
BERT-PHQ9      & 94.4 & 87.2 (0.47) & 91.8 & & 95.2 & 88.4 (0.48) & 91.8 & & 95.2 & 88.2 (0.48) & 91.4 & & 95.1 & 88.9 (0.46) & 92.5\\ 
BERT-MHS   & \underline{94.9} & \underline{88.6 (0.29)} & 93.0 & & \underline{95.4} & 89.2 (0.42) & 92.3 & & 95.7 & 90.3 (0.38) & \underline{93.7} & & \underline{95.7} & 90.0 (0.28) & \underline{93.7}\\\hline
RoBERTa          & 94.8 & \underline{88.6 (0.34)} & \underline{93.1} & & \underline{95.4} & \underline{89.4 (0.56)} & \underline{92.9} & & \underline{95.8} & \underline{90.4 (0.35)} & \underline{93.7} & & \underline{95.7} & \underline{90.3 (0.35)} & \underline{93.7}\\
RoBERTa-PHQ9     & \underline{94.9} & \underline{88.6 (0.50)} & 92.6 & & \underline{95.4} & \underline{89.4 (0.59)} & 92.6 & & 95.5 & 89.4 (0.33) & 92.4 & & 95.6 & 89.9 (0.47) & 93.3\\ 
RoBERTa-MHS  & \bf 95.5 & \bf 89.6 (0.31)* & \bf 93.8 & & \bf 95.8 & \bf 90.4 (0.31)* & \bf 93.4 & & \bf 96.2 & \bf 91.5 (0.28)* & \bf 94.3 & & \bf 95.9 & \bf 90.8 (0.26)* & \bf 94.0 \\\hline                        
\end{tabular}
\caption{
Results on four mental disorder detection tasks. Each result is averaged after 30 runs. The best results for each task are shown in bold, and the second-best results are underlined. * denotes that the performance gain is statistically significant with $p<0.05$ under all pairwise $t$-tests.
}\label{tab:result}
\vspace{-0.05in}
\end{table*}

Table~\ref{tab:result} shows the overall performance of our proposed model (MHS) and strong baselines on four tasks.
Each task is about detecting texts with corresponding mental illness episodes on social media.
We see that our model outperforms all competing approaches, including linguistic feature-based models, end-to-end pre-trained models, and a method that uses symptom-related knowledge.

Linguistic feature-based models exhibit significant performance variations based on the level of detail in their feature design.
By contrast, MHS can simply find the features directly from the contextualized representation, giving better performance improvements.
Pre-trained models with contextualized embeddings have the benefits that can be easily fine-tuned for a wide range of tasks.
However, compared to MHS, they lack a specific focus on domain-based features, while MHS is tailored to identify such features, leading to better performance.

We implemented our model and PHQ9 model with two different encoders, BERT and RoBERTa, and the tendency for performance improvement is the same on both encoders.
Both PHQ9 and MHS leverage symptom-related information but differ in their architecture, specifically whether it is a multi-step pipeline or an end-to-end model.
The end-to-end design of MHS allows for direct learning of complex relationships, reducing the potential for error propagation, and resulting in enhanced performance compared to the pipeline model.
Moreover, for this pipeline model to apply to other mental disorders, a symptom pattern must be created for each mental disorder, which is challenging to achieve without expert-level knowledge.
On the other hand, our proposed model overcomes these challenges by simply replacing symptom descriptions.
A detailed analysis of the performance improvement is shown in Section \ref{sec:analysis}.

\subsection{Model Parameters}
\begin{table}[t]
\centering
\small
\begin{tabular}{l|cc}
\hline
\makecell[c]{\bf Model}         & \bf \#parameters & \bf Relative Size \\ \hline
BERT           & 108,311,810  & 1.00\\
MHS w/bert    & 108,967,319  & 1.01\\
RoBERTa        & 124,647,170  & 1.15\\
MHS w/roberta & 125,302,679  & 1.16 \\ \hline
\end{tabular}
\caption{The numbers of parameters for BERT, RoBERTa, and our models.}\label{tab:param}
\vspace{-0.1in}
\end{table}
Table \ref{tab:param} shows the number of parameters for each model. 
Compared to the baseline models, the additional number of parameters for our siamese network is about 655K.
It is a much smaller number than that of the additional parameters for RoBERTa and BERT (about 16M), but the performance of MHS (w/bert) is slightly better or shows little difference. 
It suggests that our proposed model, learning domain knowledge, achieves much efficient performance improvement by adding just a small number of parameters.


\section{Model Analysis and Discussions}
\label{sec:analysis}

\subsection{Ablation Study}
\label{subsec:ablation}

\begin{table}[t]
\centering
\scriptsize
\vspace{0.05in}
\vspace{-0.05in}
\setlength\tabcolsep{4pt}
\setlength\extrarowheight{1pt}
  \begin{tabular}{l|ccccc}
    \hline
    \bf \makecell[c]{Model} & \bf Acc. & \bf Pre. & \bf Rec. & \bf F1 & \bf AUC\\
    \hline
    CNNs\hspace{1mm}w/bert emb. & 94.0 & \bf 89.8 & 82.9 & 86.2 & 90.1\\
    \hspace{2mm}+single-head & 94.5 & 88.6 & 86.8 & 87.6 & 91.7\\
    \hspace{2mm}+multi-head\hspace{1mm}+one description & 94.9 & 87.3 & 90.2 & 88.7  & 93.2\\
    \hspace{2mm}+multi-head\hspace{1mm}+multi-description & \bf 95.4 & 89.1 & \bf 90.5 & \bf 89.7 & \bf 93.9\\
    \hline
  \end{tabular}
  \caption{\label{tab:ablation} An ablation study of different levels of knowledge and features affecting our model. The result is the average of the four tasks.}
  \vspace{-0.15in}
\end{table}

We conducted an ablation study to investigate the effectiveness of each part in our proposed model.
We removed the siamese network from our proposed methods, resulting in just convolutional neural networks (CNNs). 
We implemented a single-head siamese network in which all sentences from all heads are put together into just one head. 
We also implemented two versions of a multi-head siamese network employing just one description or multiple descriptions, respectively.
\par
The experimental results are shown in Table~\ref{tab:ablation}.
The result shows that our proposed model gives the best performance when all of the modules are combined. 
Compared to CNN models, the performances are improved when the siamese network is added.
Note that the siamese network contributes to accurate detection, since it captures the symptom-related features by comparing target texts with symptom descriptions.
In addition, the performances are also improved when employing a multi-head rather than a single-head.
It implies that individually training each symptom yields better results than training all symptoms together, as each symptom has unique features.
Compared to learning from only one description per head, the performance of learning from multiple descriptions is improved.
It may be due to each head learning further about the symptom through various sentences, covering distinct aspects of each symptom.
\begin{table}[t]
\scriptsize
\centering
\setlength\tabcolsep{4pt}
\setlength\extrarowheight{2pt}
\begin{tabular}{l|cccccccc}
\hline
                           & \multicolumn{8}{c}{\bf Detection Task}                                                                                                                                                              \\ \cline{2-9} 
\multicolumn{1}{c|}{\bf Model} & \multicolumn{2}{c|}{\bf MDD}                           & \multicolumn{2}{c|}{\bf Bipolar}                       & \multicolumn{2}{c|}{\bf GAD}                       & \multicolumn{2}{c}{\bf BPD}       \\ \hline
\multicolumn{1}{l|}{MHS}   & F1            & \multicolumn{1}{c|}{AUC}           & F1            & \multicolumn{1}{c|}{AUC}           & F1            & \multicolumn{1}{c|}{AUC}           & F1            & AUC           \\ \hline
w/depression               & \textbf{89.6} & \multicolumn{1}{c|}{\textbf{93.8}} & 89.4          & \multicolumn{1}{c|}{92.7}          & 89.5          & \multicolumn{1}{c|}{93.4}          & 89.8          & 93.5          \\
w/bipolar                  & 88.2          & \multicolumn{1}{c|}{92.4}          & \textbf{90.4} & \multicolumn{1}{c|}{\textbf{93.4}} & 90.4          & \multicolumn{1}{c|}{93.2}          & 88.8          & 91.8          \\
w/anxiety                  & 88.5          & \multicolumn{1}{c|}{92.7}          & 89.2          & \multicolumn{1}{c|}{93.2}          & \textbf{91.5} & \multicolumn{1}{c|}{\textbf{94.3}} & 88.9          & 92.9          \\
w/bpd                      & 88.3          & \multicolumn{1}{c|}{92.4}          & 89.3          & \multicolumn{1}{c|}{92.5}          & 88.8          & \multicolumn{1}{c|}{92.8}          & \textbf{90.8} & \textbf{94.0} \\ \hline
\end{tabular}
\caption{\label{tab:cross-symptom} The results of four mental illness detection tasks. The notation \textit{w/(mental illness)} indicates the model takes symptom descriptions of the specific \textit{mental illness} as input, respectively.}
\vspace{-0.13in}
\end{table}

\begin{figure*}[ht]
\centering
\includegraphics[width=1.0\linewidth]{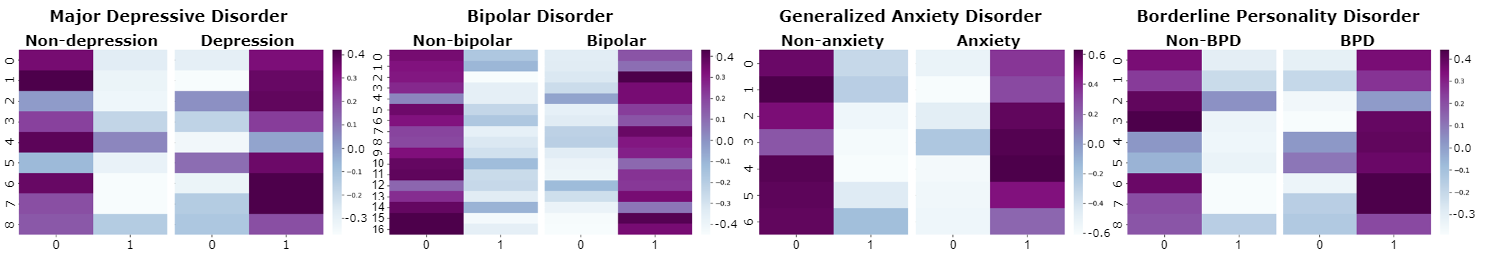}
\vspace{-0.1in}
\caption{Examples of weights learned during the training process for each task. Each row represents a distance computed by each head, indicating the particular knowledge of the related symptoms.}
\vspace{-0.1in}
\label{fig:weights}
\end{figure*}

\vspace{-0.15in}
\subsection{Contribution of Symptom Descriptions}
\label{subsec:cross-symptom}

To assess the effectiveness of symptom descriptions in detecting the presence of symptoms, we measure their performance by replacing the descriptions of symptoms with those of other mental disorders.
The results are shown in Table~\ref{tab:cross-symptom}.
We carried out four mental disorder detection tasks using four models, each utilizing symptom descriptions of four distinct mental disorders as inputs.


The models exhibit optimal performance when the input symptom description corresponds to the target mental disorder.
It suggests that, by providing the model with accurate and appropriate symptom descriptions, MHS can learn effectively to identify the specific features associated with a particular mental disorder.
This also implies that MHS can identify and utilize the nuanced distinctions in the characteristics of each symptom, leading to enhanced performance in detection.

\subsection{Cross-domain Test}
\label{subsec:cross-domain}
\begin{table}[t]
\scriptsize
\centering
\setlength\extrarowheight{1pt}
\begin{tabular}{cl|rrrr}
\hline
\multicolumn{1}{l}{}                                                                                & \multicolumn{1}{l|}{} & \multicolumn{2}{c}{\bf RSDD}                                    & \multicolumn{2}{c}{\bf eRisk}                        \\ \cline{3-6}
\multicolumn{2}{c|}{Model}                                                                                                  & \multicolumn{1}{c}{F1} & \multicolumn{1}{c|}{AUC}           & \multicolumn{1}{c}{F1} & \multicolumn{1}{c}{AUC} \\ \hline
\multicolumn{1}{c|}{\multirow{6}{*}{\begin{tabular}[c]{@{}c@{}}Train:\\ r/depression\end{tabular}}} & BERT                  & 35.7                   & \multicolumn{1}{r|}{50.8}          & 52.3                   & 78.1                    \\
\multicolumn{1}{c|}{}                                                                               & XLNet                 & 34.9                   & \multicolumn{1}{r|}{50.5}          & 52.8                   & 78.5                    \\
\multicolumn{1}{c|}{}                                                                               & RoBERTa               & 37.4                   & \multicolumn{1}{r|}{51.6}          & 52.5                   & 78.3                    \\
\multicolumn{1}{c|}{}                                                                               & GPT-2                 &  \underline{37.8}                   & \multicolumn{1}{r|}{\underline{51.7}}          & 53.2                   & 78.4                    \\
\multicolumn{1}{c|}{}                                                                               & PHQ9                  & 37.2                   & \multicolumn{1}{r|}{51.5}          & \underline{53.3}                   & \underline{78.8}                    \\
\multicolumn{1}{c|}{}                                                                               & MHS              & \textbf{38.6}          & \multicolumn{1}{r|}{\textbf{52.0}} & \textbf{54.9}          & \textbf{79.5}           \\ \hline
\end{tabular}
\caption{\label{tab:cross-dataset} The results of evaluation across the other dataset. Due to the uneven distribution of data, we report the weighted F1 scores for each test.}

\end{table}

In order to investigate the flexibility of MHS, we evaluated its performance across datasets and other mental disorders.

\par \noindent \textbf{Dataset Transferability}
Given that the ability to generalize to new and unseen data platforms is a crucial aspect of mental illness detection models \cite{harrigian2020models}, we evaluate their performance across different datasets.
We selected two datasets, RSDD~\cite{yates2017depression} and eRisk2018~\cite{losada2019overview}, to evaluate cross-dataset transfer.
Unlike our Reddit dataset (Subsection~\ref{subsec:dataset}), sourced from communities specific to certain mental illnesses, RSDD and eRisk2018 data are based on user self-reports, resulting in data that is different from and potentially unseen by the Reddit dataset.
We trained each model using the Reddit train dataset and evaluated its performance on the test sets of RSDD and eRisk2018, respectively.

As shown in Table~\ref{tab:cross-dataset}, MHS outperforms all strong baselines over all datasets. 
The improved performance of MHS compared to GPT-2, a strong few-shot learner, is likely due to its ability to leverage domain-specific knowledge.
The higher generalizability of MHS compared to PHQ9 is likely attributed to its end-to-end architecture, which allows for direct learning of symptom features from data, as opposed to PHQ9's reliance on pre-defined symptom patterns.

\par \noindent \textbf{Domain Transferability}
\begin{table}[t]
\scriptsize
\centering
\setlength\tabcolsep{4pt}
\setlength\extrarowheight{2pt}
\begin{tabular}{cl|cccccc}
\hline
\multicolumn{1}{l}{}                                                                       &                       & \multicolumn{6}{c}{\textbf{Test: Target}}                                                                                                  \\ \cline{3-8} 
\multicolumn{1}{l}{}                                                                       & \multicolumn{1}{c|}{} & \multicolumn{2}{c|}{\textbf{Bipolar}}              & \multicolumn{2}{c|}{\textbf{GAD}}              & \multicolumn{2}{c}{\textbf{BPD}} \\\cline{3-8}
\multicolumn{2}{c|}{Model}                                                                                         & F1            & \multicolumn{1}{c|}{AUC}           & F1            & \multicolumn{1}{c|}{AUC}           & F1              & AUC            \\ \hline
\multicolumn{1}{c|}{\multirow{7}{*}{\begin{tabular}[c]{@{}c@{}}Train:\\ MDD\end{tabular}}} & Feature               & 54.0          & \multicolumn{1}{c|}{69.1}          &    49.5           & \multicolumn{1}{c|}{66.6}              &     55.2            &       69.8         \\
\multicolumn{1}{c|}{}                                                                      & BERT                  & 62.0          & \multicolumn{1}{c|}{73.7}          & 51.7          & \multicolumn{1}{c|}{67.8}          & 60.9            & 72.8           \\
\multicolumn{1}{c|}{}                                                                      & XLNet                 & 65.2          & \multicolumn{1}{c|}{75.4}          & 51.3          & \multicolumn{1}{c|}{67.6}          & 60.5            & 72.6           \\
\multicolumn{1}{c|}{}                                                                      & RoBERTa               & 65.1          & \multicolumn{1}{c|}{75.6}          & {\underline{58.6}}    & \multicolumn{1}{c|}{71.6}          & \underline{64.9}      & \underline{75.4}     \\
\multicolumn{1}{c|}{}                                                                      & GPT-2                 & 65.2          & \multicolumn{1}{c|}{75.7}          & \textbf{59.6} & \multicolumn{1}{c|}{\underline{72.1}}    & 62.6            & 73.5           \\ \cline{2-8} 
\multicolumn{1}{c|}{}                                                                      & MHS w/depression             & {\underline{66.7}}    & \multicolumn{1}{c|}{{\underline{76.6}}}    & 55.5          & \multicolumn{1}{c|}{69.8}          & 60.2            & 72.6           \\
\multicolumn{1}{c|}{}                                                                      & MHS w/($=$Target)        & \textbf{76.6} & \multicolumn{1}{c|}{\textbf{85.4}} & \textbf{59.6} & \multicolumn{1}{c|}{\textbf{72.2}} & \textbf{67.5}   & \textbf{77.3}  \\ \hline
\end{tabular}
\caption{\label{tab:cross-disorder} The results of evaluation across the other mental disorders.}
\vspace{-0.15in}
\end{table}
As suggested by some researchers \cite{aich2022really, uban2022multi}, we evaluated the transferability of MHS across other mental disorders by training the models on a depression dataset and testing on other mental disorder datasets (see Table~\ref{tab:cross-disorder}). 
The results of the experiments indicate that MHS significantly outperforms all relevant baselines, particularly when it utilizes symptoms that match the target mental disorder.
This suggests that the transferability of the model can be significantly enhanced by simply replacing symptom descriptions.
This also implies that it may be feasible to develop a model that can classify texts related to various other mental disorders if the symptoms of those disorders are provided appropriately.

\subsection{Interpretation}
\label{subsec:interpretation}

Using our model, we can interpret the detected results by analyzing their representations of learned weights and distance values.
In order to see if our model properly learned symptom-related knowledge from a few descriptions and identified similar stories from the target texts, we looked into the learned weights produced by the last step of our model, the fully connected layer.
To show the effectiveness of MHS, we visualize the examples of learned weights from training steps in Figure~\ref{fig:weights}.
The color scale represents the strength of the learned weights (i.e., the distance values of each head).
Each row represents heads, indicating each symptom referring to Table~\ref{tab:symptoms}, and each column represents the labels. 
We observe a clearly contrasting pattern in the distance weights for each task.

We could also identify which symptoms are mainly activated or not by investigating the learned weights during the training process.
For example, in detecting MDD-related texts, most of the symptoms have higher weights than depression. 
It suggests that most of the symptoms give rise to a major role during the detection process. 
\begin{figure}
\centering
\includegraphics[width=\linewidth]{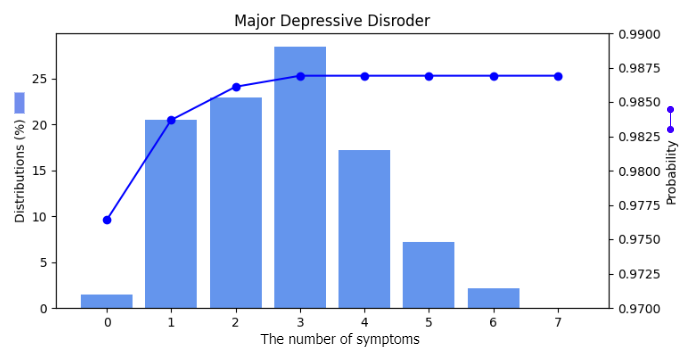}
\caption{The number of salient symptoms and probability of the final output from true-positive samples in MDD detection.}
\label{fig:num_sam}
\vspace{-0.2in}
\end{figure}

An important criterion in diagnosing a mental illness by experts is the number of expressed symptoms.
The number of symptoms must exceed a certain number to be diagnosed as a corresponding mental illness.
In order to see if the human-level diagnostic process works in our model as well, we looked into the number of salient symptoms in true-positive samples.
We calculated percentiles from the similarity scores for each symptom in the true-positive samples from test sets, and set the threshold by 70\% of the percentile.
Then, when exceeding the threshold set by the criterion, the symptom was selected as a prominent feature in the text.
We present the distribution of the numbers of salient symptoms and their averaged probabilities of the final output from test sets of detecting MDD-related texts in Figure \ref{fig:num_sam}.

In our model, the average probability is relatively low when there are fewer than three symptoms, but for three symptoms or more, our model makes a decision with high confidence at a similar level. 
It suggests that MHS also detects mental disorder-related texts with high confidence when the number of symptoms exceeds a specific number, the same as when humans diagnose.
The criterion number being smaller in MHS may be due to the shorter length of social media texts, which may not fully convey the user's background and lifestyle.

\subsection{Case Study}
\begin{table*}[t]
\scriptsize
\centering
\begin{tabular}{c|l|l}
\hline
\textbf{No.}                                                & \multicolumn{1}{c|}{\textbf{Example}}                                                                                                                                                                                                                                                                                                                                 & \multicolumn{1}{c}{\textbf{Expected Symptoms}}                                                                                                \\ \hline
\textbf{\begin{tabular}[c]{@{}c@{}}1.\\ (MDD)\end{tabular}} & \begin{tabular}[c]{@{}l@{}}Whenever I wake up in the morning, I hate myself, and I want to commit suicide. I didn't\\ have any friends to hang out with because I did not need to make friends actively when I went\\  to school. The only reason I am not committing suicide is I don't want my parents to cry.\end{tabular}                                         & \begin{tabular}[c]{@{}l@{}}\textbf{D0 (81\%)} \textbf{D1 (80\%)} \textcolor{lightgray}{D2 (25\%)}\\ \textcolor{lightgray}{D3 (56\%)} \textcolor{lightgray}{D4 (10\%)} \textcolor{lightgray}{D5 (40\%)}\\ \textcolor{lightgray}{D6 (47\%)} \textcolor{lightgray}{D7 (61\%)} \textbf{D8 (71\%)}\end{tabular} \\ \hline
\textbf{\begin{tabular}[c]{@{}c@{}}2.\\ (GAD)\end{tabular}} & \begin{tabular}[c]{@{}l@{}}I often feel anxious that something terrible is about to happen. For example, my husband\\ will likely lose his job, or a family member will become ill or have an accident.\\ I know these worries are unnecessary and excessive, but I can't stop worrying.\\ I'm always nervous, so I feel exhausted even if I do nothing.\end{tabular} & \begin{tabular}[c]{@{}l@{}}\textcolor{lightgray}{A0 (41\%)} \textbf{A1 (72\%)} \textbf{A2 (75\%)}\\ \textbf{A3 (79\%)} \textcolor{lightgray}{A4 (48\%)} \textcolor{lightgray}{A5 (34\%)}\\ \textcolor{lightgray}{A6 (37\%)}\end{tabular}                       \\ \hline
\end{tabular}
\caption{
Examples of texts related to MDD and GAD, respectively, and the corresponding symptoms that the models provide for interpretation. The notations of each symptom are referred to in Table~\ref{tab:symptoms}.
}\label{tab:case-study}
\vspace{-0.2in}
\end{table*}

For the case study, we made an example based on the samples corresponding to each mental disorder in the psychology major textbook.
We present example sentences for MDD and GAD (Table~\ref{tab:case-study}), and the model's predictions were correct in both cases.
We set the same threshold as shown in Figure~\ref{fig:num_sam}.
The dominant symptoms predicted by the model are D0 (\textit{depressed mood}), D1 (\textit{diminished interest}), and D8 (\textit{suicidal ideation}), for MDD, and A1 (\textit{difficult to control the worry}), A2 (\textit{irritability}), and A3 (\textit{easily fatigued}), for GAD.
In the case of D0 and D1 in MDD, our model captures the feature related to the symptom, despite the absence of the term `\textit{depress}' or `\textit{interest}'.
These cases support the assumption that our model can detect and interpret when symptoms of a particular mental illness are prominent in text.



\section{Conclusion}
\label{sec:conclusion}

In this paper, we proposed a simple but flexible model for detecting texts containing contents of mental health problems.
Our model outperformed the state-of-the-art models and achieved human-interpretable results over symptoms regarding mental disorders.
The proposed model demonstrates an exceptional ability to utilize domain knowledge as it is designed to capture relevant features from texts directly.
Experimental results also indicate that MHS can quickly adapt to other mental disorder domains by simply replacing symptom descriptions.
The scope of this paper was limited to the investigation of four mental disorder detection tasks.
Nevertheless, this approach can be extended to other mental health conditions as long as the symptom-relevant questionnaires are provided accordingly.

\section*{Limitations}
It should be noted that, as our model and the baseline models in this study were trained using texts from social media and the experiments were conducted on online text, the results may not accurately reflect the performance in a clinical setting.
A proper diagnosis by clinical experts necessitates a comprehensive analysis of various factors, including the number of manifested symptoms, the onset and history of symptoms, developmental background, lifestyle, and recent life changes, in order to gain a comprehensive understanding of the patient's condition.
However, it is still challenging to capture detailed information such as personal secrets through online text, as these texts are often composed of fragments of daily life, episodic experiences, and emotive expressions rather than providing a comprehensive view of an individual's life.
Despite the domain-specific limitations imposed by the fragmentary text, we hope that our model may still serve as a valuable aid for clinical experts in their decision-making process.
Furthermore, future research should aim to move beyond predicting psychological symptoms and disorders solely based on linguistic styles and expressions, and instead seek to uncover the underlying features that contribute to these expressions as our model does.

\section*{Ethics Statement}
Since privacy concerns and the risk to the individuals should always be considered, especially using social media data, we have employed mechanisms to avoid any harmful and negative consequences of releasing our model. To this end, we removed individually identifiable information such as user names, user IDs, or e-mail addresses. We also removed any URLs from our data not to be trained on such personal information in our model.
As for the use of open datasets in this work, we used them in accordance with guidelines that allow their use within the established usage policy.
Especially we ensure that no attempts can be made to establish contact with specific individuals or deanonymize users in the datasets.

Our paper may contain direct references to specific disorders or diseases (such as psychiatric patients, Siamese, or names of mental disorders) and expressions that could be considered offensive to particular individuals.
We want to emphasize that these expressions are used solely for the purpose of academic discourse and are not intended to be disrespectful or offend anyone.

In addition, our proposed model is not intended to label or stigmatize individuals online but rather to serve as a warning system for potential threats to personal well-being and public health.
It is important to note that even if this model identifies potential mental illnesses and symptoms, it should not be considered a definitive diagnosis.
Still, the model provides an indication of the likelihood of a disorder; it should be used as a reference for self-diagnose and in consultation with a mental health expert for an official diagnosis.
An official diagnosis and results require consultation with medical and psychological experts, and this system aims at serving as an aid in the diagnostic process.
We make our implementation code publicly available for research purposes, and we hope it will be used to improve the lives of individuals suffering from mental illnesses.

\section*{Acknowledgements}
This work was supported by the National Research Foundation of Korea (NRF) (No. RS-2023-00208054, A multi-modal abusive language detection system and automatic feedback with correction) grant funded by the Korean government.

\bibliography{anthology,custom}
\bibliographystyle{acl_natbib}
\newpage
\appendix

\section{Experimental Setups}
\label{appendix:experimental}

We implemented two feature-based models, support vector machine (SVM) and random forest (RF).
We fine-tuned SVM with Gaussian kernel and set \textit{C} to 100, and RF set max depth to 100.
We employed BERT's vocabulary to train \textbf{BoW} models.
For \textbf{Feature} models, we used a pre-trained sentiment classification model, and a Part-of-Speech Tagging model from the Huggingface library \cite{wolf2019huggingface}.
We fine-tuned the transformer baseline models employing the default settings from the Huggingface library:
\textbf{BERT} (\textit{bert-base-cased}), \textbf{XLNet} (\textit{xlnet-base-cased}), \textbf{RoBERTa} (\textit{roberta-base}), \textbf{GPT-2} (\textit{gpt2}). For all experiments, we set the batch size as 8 and fine-tuned all models on a single 24GB GeForce RTX 3090 GPU.
For the implementation of the \textbf{PHQ9} model, we follow the structure of the questionnaire-depression pair models by using the publicly available code from PHQ9\footnote{https://github.com/thongnt99/acl22-depression-phq9} \cite{nguyen2022improving}.
We utilized the symptom patterns which are provided by \citet{nguyen2022improving}.
We trained each of the models using all six randomly selected seeds, and all the models were trained for 3 epochs.
We optimize the model parameters of all models with the Adam optimizer \cite{kingma2014adam}.
The learning rates for BERT, XLNet, and RoBERTa models were manually fine-tuned, choosing one out of \{1e-05, 2e-05, 1e-06, 2e-06\} that shows the best F1 score.
The learning rate for GPT-2 was selected from \{1e-05, 2e-05\}, and for PHQ9, the learning rate was set to 1e-03, which was provided as an optimized hyperparameter.

\section{Comparison with Large Language Model}

Recent developments in large language models (LLMs), such as GPT-3 \cite{brown2020language}, have demonstrated strong zero-shot performance across various NLP tasks.
LLMs have the ability to achieve high performance without fine-tuning for downstream tasks, even with only zero or few examples, due to their large number of pre-trained parameters.

We experimented with obtaining results for the examples referred to in Table~\ref{tab:case-study} by using GPT-3, a widely recognized LLM.
To this end, we utilized instructional prompts by listing symptom descriptions for a specific mental illness.
The examples of prompt input and the result are shown in Table~\ref{tab:gpt-3}.
The experimental results show that the model successfully outputs the classification results in a sentence when given instructional prompts for a specific mental illness.
However, the process of selecting symptoms appears to focus on identifying multiple symptoms rather than pinpointing a specific symptom with precision.

These examples are presented for demonstration purposes only, and the results may vary depending on the utilization of different prompt optimizations \cite{liu2021gpt, qin2021learning}.
This aspect of research is beyond the scope of our current study; thus, there is room for further research to be conducted in future work.


\section{Details of Symptom Descriptions}
\label{appendix:symptom}

\begin{figure}[ht]
\centering
\includegraphics[width=0.9 \linewidth]{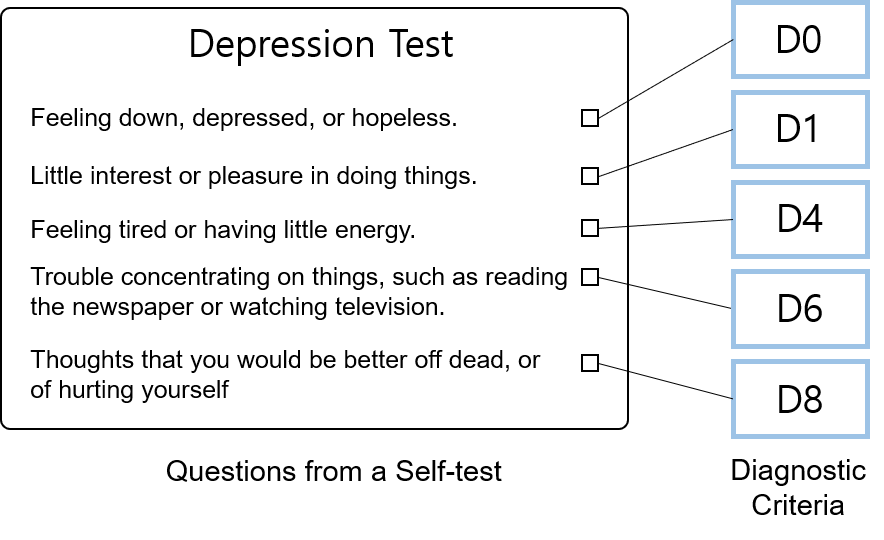}
\caption{An example mapping of questions into corresponding diagnostic criteria.}
\label{fig:symptom}
\end{figure}

In this section, we present the symptom descriptions that were utilized in our current study.
Table~\ref{tab:full-sym} shows the complete list of symptom descriptions.
We used \textit{Diagnostic and Statistical Manual of Mental Disorders (DSM-5)} \cite{american2013diagnostic} as a reference for the symptom descriptions, as it provides comprehensive guidelines for identifying symptoms of various mental disorders.
We also incorporated publicly available clinical questionnaires from online sources.
Subsequently, under the guidance of a psychology researcher, we conducted a mapping process of the questions in the self-test to the corresponding diagnostic criteria, as depicted in Figure~\ref{fig:symptom}.

\begin{table*}[t]
\scriptsize
\centering
\begin{tabular}{c|l}
\hline
\textbf{No.}                                                                 & \multicolumn{1}{c}{\textbf{Example}}                                                                                                                                                                                                                                                                                                                                                                                                                                                                                                                                                                                                                                                    \\ \hline
\multirow{3}{*}{\textbf{\begin{tabular}[c]{@{}c@{}}1.\\ (MDD)\end{tabular}}} & \begin{tabular}[c]{@{}l@{}}\textbf{Prompt:}\\ These are symptom descriptions of major depressive disorder:\\ D0. Depressed mood most of the day\\ D1. Diminished interest or pleasure\\ D2. Sleep disorders (insomnia or hypersomnia)\\ D3. Changes in weight or appetite when not dieting\\ D4. Fatigue or loss of energy\\ D5. Feeling worthlessness or guilty\\ D6. Diminished ability to think or concentrate\\ D7. A slowing down of thought and a reduction of physical movement\\ D8. Recurrent thoughts of death and suicidal ideation\\ \\ Please tell me if the user below has major depressive disorder\\ and which symptoms does the user have? (choose from the above)\end{tabular} \\ \cline{2-2} 
                                                                             & \begin{tabular}[c]{@{}l@{}}\textbf{Input:}\\ Whenever I wake up in the morning, I hate myself, and I want to commit suicide.\\ I didn't have any friends to hang out with because I did not need to make friends\\ actively when I went to school. The only reason I am not committing suicide is \\ I don't want my parents to cry.\end{tabular}                                                                                                                                                                                                                                                                                                                                                \\ \cline{2-2} 
                                                                             & \begin{tabular}[c]{@{}l@{}}\textbf{Output:}\\ Yes, this user appears to have major depressive disorder and they have symptoms\\ D0 (depressed mood most of the day), D1 (diminished interest or pleasure),\\ D5 (feeling worthlessness or guilty), D6 (diminished ability to think or concentrate), \\ and D8 (recurrent thoughts of death and suicidal ideation).\end{tabular}                                                                                                                                                                                                                                                                                                                  \\ \hline
\multirow{3}{*}{\textbf{\begin{tabular}[c]{@{}c@{}}2.\\ (GAD)\end{tabular}}} & \begin{tabular}[c]{@{}l@{}}\textbf{Prompt:}\\ These are symptom descriptions of generalized anxiety disorder:\\ A0. Excessive anxiety and worry more than 6 monts\\ A1. Difficult to control the worry\\ The anxiety and worry are associated with followings:\\ A2. Irritability\\ A3. Being easily fatigued\\ A4. Sleep disturbance\\ A5. Difficulty concentrating or mind going black\\ A6. Muscle tension\\ \\ Please tell me if the user below has generalized anxiety disorder\\ and which symptoms does the user have? (choose from the above)\end{tabular}                                                                                                                               \\ \cline{2-2} 
                                                                             & \begin{tabular}[c]{@{}l@{}}\textbf{Input:}\\ I often feel anxious that something terrible is about to happen. \\ For example, my husband will likely lose his job, or a family member will become ill\\  or have an accident. I know these worries are unnecessary and excessive, but \\ I can't stop worrying. I'm always nervous, so I feel exhausted even if I do nothing.\end{tabular}                                                                                                                                                                                                                                                                                                       \\ \cline{2-2} 
                                                                             & \begin{tabular}[c]{@{}l@{}}\textbf{Output:}\\ Yes, the user has generalized anxiety disorder and they have the following symptoms:\\  A0. Excessive anxiety and worry more than 6 months, A1. Difficult to control the worry,\\  A2. Irritability, A3. Being easily fatigued, A4. Sleep disturbance, \\ A5. Difficulty concentrating or mind going blank, A6. Muscle tension.\end{tabular}                                                                                                                                                                                                                                                                                                       \\ \hline
\end{tabular}
\caption{\label{tab:gpt-3} Results for input example texts related to MDD and GAD, using GPT-3, respectively.}
\end{table*}
\begin{table*}[ht]
\centering
\begin{subtable}{\textwidth}
\scriptsize
  \begin{tabular}{p{15cm}}
    \hline
        \multicolumn{1}{c}{\bf Major Depresive Disorder}  \\
        \makecell[l]{D0: \bf Depressed mood most of the day, nearly every day.\\
        \hspace{3.5mm} \underline{Feeling down, depressed, or hopeless.}\\
        D1: \bf Markedly diminished interest or pleasure in all, or almost all, activities most of the day, nearly every day.\\
        \hspace{3.5mm} \underline{Little interest or pleasure in doing things.}\\
        D2: \bf Insomnia or hypersomnia nearly every day.\\
        \hspace{3.5mm} \underline{Trouble falling or staying asleep, or sleeping too much.}\\
        D3: \bf Significant weight loss when not dieting or weight gain, or decrease or increase in appetite nearly every day.\\
        \hspace{3.5mm} \underline{Poor appetite or overeating.}\\
        D4: \bf Fatigue or loss of energy nearly every day.\\
        \hspace{3.5mm} \underline{Feeling tired or having little energy.}\\
        D5: \bf Feeling worthlessness or excessive or inappropriate guilt nearly every day.\\
        \hspace{3.5mm} \underline{Feeling bad about yourself - or that you are a failure or have let yourself or your family down.}\\
        D6: \bf Diminished ability to think or concentrate, or indecisiveness, nearly every day.\\
        \hspace{3.5mm} \underline{Trouble concentrating on things, such as reading the newspaper or watching television.}\\
        D7: \bf A slowing down of thought and a reduction of physical movement.\\
        \hspace{3.5mm} \underline{Moving or speaking so slowly that other people could have noticed.}\\
        D8: \bf Recurrent thoughts of death, recurrent suicidal ideation without a specific plan, or a suicide attempt or a specific plan for committing suicide.\\
        \hspace{3.5mm} \underline{Thoughts that you would be better off dead, or of hurting yourself.}\\
        } \\
        \hline
      \end{tabular}
  \end{subtable}
  \begin{subtable}{\textwidth}
\scriptsize
  \begin{tabular}{p{15cm}}
       \multicolumn{1}{c}{\bf Bipolar Disorder}  \\
        \makecell[l]{
        \bf Major Depressive Episode: \hspace{1mm}D0-D8: Same as major depressive disorder.\\
        \bf Manic Episode:\\
        M0: \bf A distinct period of abnormally and persistently elevated, expansive, or irritable mood and abnormally and persistently increased \\ \bf goal-directed activity or energy, lasting at least 1 week and present most of the day, nearly every day.\\
        \hspace{3.5mm} \underline{Do you ever experience a persistent elevated or irritable mood for more than a week?}\\
        M1: \bf Increase in goal-directed activity or psychomotor agitation (i.e., purposeless non-goal-directed activity).\\
        \hspace{3.5mm} \underline{Do you ever experience persistently increased goal-directed activity for more than a week?}\\
        M2: \bf Inflated self-esteem or grandiosity.\\
        \hspace{3.5mm} \underline{Do you ever experience inflated self-esteem or grandiose thoughts about yourself?}\\
        M3: \bf Decreased need for sleep (e.g., feels rested after only 3 hours of sleep).\\
        \hspace{3.5mm} \underline{Do you ever feel little need for sleep, feeling rested after only a few hours?}\\
        M4: \bf More talkative than usual or pressure to keep talking.\\
        \hspace{3.5mm} \underline{Do you ever find yourself more talkative than usual?}\\
        M5: \bf Flight of ideas or subjective experience that thoughts are racing.\\
        \hspace{3.5mm} \underline{Do you experience racing thoughts or a flight of ideas?}\\
        M6: \bf Distractibility (i.e., attention too easily drawn to unimportant or irrelevant external stimuli), as reported or observed.\\
        \hspace{3.5mm} \underline{Do you notice (or others comment) that you are easily distracted?}\\
        M7: \bf Excessive involvement in activities that have a high potential for painful consequences.\\
        \hspace{3.5mm} \underline{Do you engage excessively in risky behaviors, sexually or financially?}\\
        } \\
        \hline
      \end{tabular}
  \end{subtable}
  \begin{subtable}{\textwidth}
\scriptsize
  \begin{tabular}{p{15cm}}
        \multicolumn{1}{c}{\bf Anxiety Disorder}  \\
        \makecell[l]{A0: \bf Excessive anxiety and worry, occurring more days than not for at least 6 months, about a number of events or activities.\\
        \hspace{3.5mm} \underline{Do you worry about lots of different things?}
        \hspace{3.5mm} \underline{Do you worry about things working out in the future?}\\
        \hspace{3.5mm} \underline{Do you worry about things that have already happened in the past?}
        \hspace{3.5mm} \underline{Do you worry about how well you do things?}\\
        A1: \bf The individual finds it difficult to control the worry.\\
        \hspace{3.5mm} \underline{Do you have trouble controlling your worries?}
        \hspace{3.5mm} \underline{Do you feel jumpy?}\\
        A2: \bf The anxiety and worry are associated with irritability.\\
        \hspace{3.5mm} \underline{Do you get irritable and/or easily annoyed when anxious?}\\
        A3: \bf The anxiety and worry are associated with being easily fatigued.\\
        \hspace{3.5mm} \underline{Does worry or anxiety make you feel fatigued or worn out?}\\
        A4: \bf The anxiety and worry are associated with sleep disturbance (difficulty falling or staying asleep, or restless, unsatisfying sleep).\\
        \hspace{3.5mm} \underline{Does worry or anxiety interfere with falling or staying asleep?}\\
        A5: \bf The anxiety and worry are associated with difficulty concentrating or mind going blank.\\
        \hspace{3.5mm} \underline{Does worry or anxiety make it hard to concentrate?}\\
        A6: \bf The anxiety and worry are associated with muscle tension.\\
        \hspace{3.5mm} \underline{Do your muscles get tense when you are worried or anxious?}\\
        } \\
        \hline
      \end{tabular}
  \end{subtable}
  \begin{subtable}{\textwidth}
\scriptsize
  \begin{tabular}{p{15cm}}
        \multicolumn{1}{c}{\bf Borderline Personality Disorder}  \\
        B0: \bf A pattern of unstable and intense interpersonal relationships characterized by alternating between extremes of idealization and devaluation.\\
        \hspace{3mm} \underline{My relationships are very intense, unstable, and alternate between the extremes of over idealizing and undervaluing people who are important to me.}\\
        B1: \bf Recurrent suicidal behavior, gestures, or threats, or self-mutilating behavior.\\
        \hspace{3.5mm} \underline{Now, or in the past, when upset, I have engaged in recurrent suicidal behaviors, gestures, threats, or self-injurious behavior} \\ \hspace{3.5mm} \underline{such as cutting, burning, or hitting myself.}\\
        B2: \bf Identity disturbance: markedly and persistently unstable self-image or sense of self.\\
        \hspace{3.5mm} \underline{I have a significant and persistently unstable image or sense of myself, or of who I am or what I truly believe in.}\\
        B3: \bf Affective instability due to a marked reactivity of mood.\\
        \hspace{3.5mm} \underline{My emotions change very quickly, and I experience intense episodes of sadness, irritability, and anxiety or panic attacks.}\\
        B4: \bf Inappropriate, intense anger or difficulty controlling anger.\\
        \hspace{3.5mm} \underline{My level of anger is often inappropriate, intense, and difficult to control.}\\
        B5: \bf Transient, stress-related paranoid ideation or severe dissociative symptoms.\\
        \hspace{3.5mm} \underline{I have very suspicious ideas, and am even paranoid or I experience episodes under stress when I feel that I, other people,} \\\hspace{3.5mm}\underline{or the situation is somewhat unreal.}\\
        B6: \bf Impulsively in at least two areas that are potentially self-damaging (e.g., spending, sex, substance abuse, reckless driving, binge eating).\\
        \hspace{3.5mm} \underline{I engage in two or more self-damaging acts such as excessive spending, unsafe and inappropriate sexual conduct, substance abuse,} \\\hspace{3.5mm}\underline{reckless driving, and binge eating.}\\
        B7: \bf Frantic efforts to avoid real or imagined abandonment.\\
        \hspace{3.5mm} \underline{I engage in frantic efforts to avoid real or imagined abandonment by people who are close to me.}\\
        B8: \bf Chronic feelings of emptiness.\\
        \hspace{3.5mm} \underline{I suffer from feelings of emptiness and boredom.}\\
         \\
        \hline
      \end{tabular}
  \end{subtable}
  \vspace{-0.1in}
  \caption{\label{tab:full-sym} The complete list of collected sentences for each head. The diagnostic criteria, sourced from DSM-5, are shown in bold, and questions from clinical questionnaires are underlined.}
  \vspace{-0.1in}
\end{table*}

\end{document}